\documentclass[12pt]{article}%
\usepackage[slantedGreek]{mathpazo}
\usepackage[pdftex]{graphicx}
\usepackage{amsfonts}
\usepackage{setspace}
\usepackage{amssymb}
\usepackage{amsthm}
\usepackage{graphicx,float}
\usepackage{amsmath}%
\usepackage{subcaption}
\usepackage{color,soul,marginnote}
\usepackage{hyperref}
\usepackage{tikz}
\usepackage[round]{natbib}
\usepackage[margin=1in]{geometry}

\include{emacscomm-JD}

\title{Deep Learning for Energy Markets}

\author{
Michael Polson\thanks{Michael Polson is a researcher at Bates White. email:michael.alan.polson@gmail.com} and Vadim Sokolov\thanks{Vadim Sokolov is Assistant Professor in Operations Research at George Mason University. email:vsokolov@gmu.edu}}
\date{First Draft: April 2018\\This Draft: April 2019}
\graphicspath{{./fig/}}

\begin{document}
\maketitle
\begin{abstract}
 

\noindent Deep Learning is applied to energy markets to predict extreme loads observed in energy grids. Forecasting energy loads and prices is challenging due to sharp peaks and troughs that arise due to supply and demand fluctuations from intraday system constraints. We propose deep spatio-temporal models and extreme value theory (EVT) to capture theses effects and in particular the tail behavior of load spikes. Deep LSTM architectures with  ReLU and $\tanh$ activation functions can model trends and temporal dependencies while EVT captures highly volatile load spikes above a pre-specified threshold. To illustrate our methodology, we use hourly price and demand data from 4719 nodes of the PJM interconnection,  and we construct a deep predictor. We show that DL-EVT outperforms traditional Fourier time series methods, both in-and out-of-sample, by capturing the observed nonlinearities in prices. Finally, we conclude with directions for future research.

\vspace{0.1in} 
\end{abstract}

\singlespacing

\newpage

\section{Introduction}
Deep learning (DL) in conjunction with Extreme Value Theory (EVT) is proposed to predict load and wholesale energy prices on the energy grid. This is essential for the economic operation of grid resources, as electricity grids operate without large amounts of storage, and generation of energy (supply) within the system must always match the demand of energy (load). Electricity price prediction is challenging due to a number of complex factors that impact the intraday grid conditions, which create highly volatile price spikes. Deep Learning multi-layer networks are developed to capture nonlinearities and the spatio-temporal patterns in energy prices and demand. Likelihood function defined by Extreme Value Theory allows to properly model spikes. As supply must constantly adapt to meet changes in load, accurate predictions are essential to making informed short and long-term generation decisions. Accurate anticipation of fluctuations in load, especially sharp fluctuations, would remove certain flexibility constraints, allowing for efficient deployment of generation and grid resources. 

The traditional approach to electricity price prediction has been applying economic models based on firm behavior to the data. More recently, data-driven analytics, using large price data sets and machine learning techniques have been used to uncover price patterns. Forecasting supply and demand with a standard deep learning model, however, fails to address the importance of peak prediction. Deep learning models aim to predict the mean level of the dependent variable and typically do not capture any extreme jumps in the data. Furthermore, squared loss is used to fit the model, which implicitly assumes the Gaussian distribution of the errors. Therefore, in the context of electricity markets, a Gaussian model would be well suited for predicting the system's average demand or energy price, but would not capture the peaks nor the true, fat-tailed distribution of the dependent variables. Until now, data-driven models were not flexible enough to capture the extreme nonlinearities in the price the dynamics. Recently, deep learners (DL) have shown empirical success in large datasets forecasting problems with high dimensional nonlinearities. Long-short-term memory (LSTM) provides a framework for building spatio-temporal model \citep{dixon2017deep,polson2017}.

Modeling extremes has a long history in financial risk management~\cite{poon2003extreme}. For example, \cite{hilal2011hedging} develop an extreme value theory model for stock indexes. Like predicting price in equity markets, predicting price in wholesale electricity markets is challenging due to a number of complex factors that impact intraday supply and demand. 

\cite{davison2012statistical} develop a spatial statistical model for the extremes of a natural process. Peaks are modeled as an exceedance of a certain threshold. EVT provides the framework for the prediction of these exceedances, it predicts the frequency of energy price exceeding a certain threshold~\cite{davison1990models}. The exceedance over a threshold allows to measure risk associated with high prices~\cite{smith2002measuring}. Incorporating extreme value theory (EVT) into deep learning allows us to capture the tail behavior of the price distribution. We develop deep learning multi-layer networks to capture nonlinearities and spatiotemporal patterns in the price distribution. In particular, likelihood functions, defined using the Extreme Value Theory framework, allow us to properly model price spikes. In the context of energy markets capturing the spikes is a crucial as these are the central component of interest in the market. Furthermore, this approach provides an improvement over traditional deep learning approaches, which typically only focus on capturing the mean of a given distribution. Our work builds on that of \cite{sigauke2013extreme} who develop probabilistic EVT model and \cite{shenoy2014risk} who use generalized linear model, with  EVT errors to model electricity demand.

The rest of our paper is organized as follows.  Section~\ref{sec:lit} provides connections to previous work. Section~\ref{sec:pred} describes the energy market for electricity and the PJM interconnection. Section~\ref{sec:dl} discusses traditional deep learning models. Section~\ref{sec:evt} extends DL models using extreme value theory (EVT). Section~\ref{sec:application} provides algorithms for load and price prediction for PJM. Finally, Section~\ref{sec:discussion} concludes with directions for future research. 

\subsection{Connection to Previous Work}\label{sec:lit}
Data-driven energy pricing models used to forecast hourly locational marginal prices (LMPs) have been studied previously ~\citep{Catalao,Hong, hong2001locational,Kim}.  \cite{hong2001locational}  propose  neural networks to predict LMPs in the PJM Interconnection. \cite{Mandal} use neural networks to improve performance, and \cite{Catalao,Kim} predict LMPs in Nord Pool, an electricity spot market located across Northern Europe. \cite{Wang} predict prices at various hubs in the American Midwest with a stacked denoising auto-encoder, exploiting local information to improve its predictive performance. Modeling wind generated-electricity is considered by \cite{hering2010powering}.  Our analysis extends the functional data analysis approach for electricity pricing developed by \cite{liebl2013modeling}.

In another line of research, \cite{cottet2003bayesian} and \cite{wilson2018quantifying} develop a random effects Bayesian framework to quantify uncertainty in wholesale electricity price projections. \cite{jonsson2013forecasting} forecast electricity prices  while accounting for wind power prediction. \cite{christensen2012forecasting} forecast spikes in electricity prices. Heavy tails in electricity prices are modeled using multivariate skew $t$-distributions by \cite{cottet2003bayesian}.  \cite{benth2014pricing} address the non-Gaussian nature of price data using a L{'e}vy process. \cite{dupuis2017electricity} develops a detrended correlation approach to capture price dynamics within the New York section of the grid. \cite{garcia2005garch} explain time-varying volatility in prices using GARCH effects for one-day price forecasting.  \cite{li2007day} develop a fuzzy inference system  to forecast prices in LMP spot markets. \cite{subbayya2013model} address the problem of model selection.  

Our approach uses extreme value theory (EVT) in combination with deep learning. To our knowledge, this is the first time EVT-based deep learning model is developed and applied. 

\section{Energy Prices in PJM Interconnection}\label{sec:pred}
The PJM Interconnection  is a regional transmission organization (RTO), that  exists to create a competitive wholesale electricity market, coordinating numerous wholesale electricity producers and consumers in all or parts of 13 states located in America's Mid-Atlantic and Great Lakes Regions as well as the District of Columbia. 

\begin{figure}[H]
	\centering
	\includegraphics[width=0.8\linewidth]{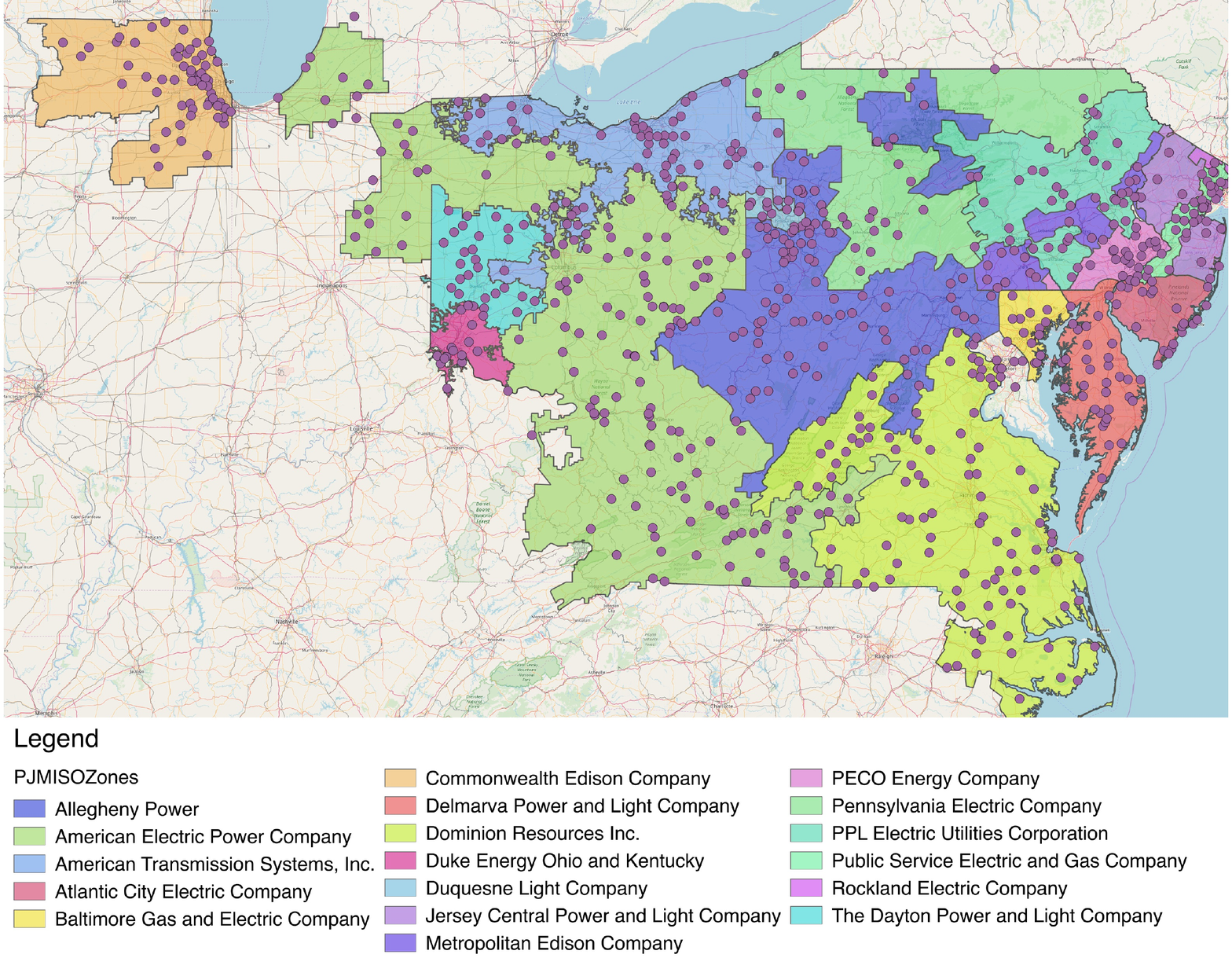}
	\caption{PJM Zone boundaries and node locations}
	\label{fig:map}
\end{figure}

PJM is divided into 20 transmission zones. Each zone is owned and operated by separate transmission owners who are responsible for designing and maintaining their portion of the system. Figure \ref{fig:map} shows PJM's load nodes and zone boundaries. Individual utilities within PJM plan their use of resources around peak loads. Predicting the strength and timing of these peaks is integral to improving both short- and long-term decision-making. Current methods used for short-term prediction focus on neural networks (weather channel, PJM). 

PJM acts as a guarantor of system reliability and is responsible for preventing outages within the system. PJM  operates the system at a cost-efficient level by coordinating generating plant operations, which are owned by various entities, to match the system's demand. Operating the system includes ensuring real-time demand is met, maintaining a reserve capacity of generation, and monitoring transmission lines to prevent overloaded lines, which could cause system failure \citep{cain2007common}.

The PJM Interconnection contains over 11,000 nodes for which hourly day-ahead or real-time prices are reported. These nodes are specific generation or load locations, aggregates of various locations, regions, or points of interconnection with areas outside of PJM. We examine prices at the 4,700 load nodes across the system.

Within the PJM Interconnection, nearly all wholesale electricity is bought or sold through bilateral contracts. The remainder is bought or sold on the two bid-based electricity markets PJM operates: day-ahead (DA) and real-time (RT) markets. In the day-ahead market, market participants submit bids or offers to buy or sell energy to the scheduling operator (PJM). The operator uses the bid and offers to determine the day-ahead LMP, which reflects the \textit{expected} cost of energy, congestion and transmission loss needed to provide electricity at a location given the \textit{expected} system constraints. 

The real-time market operates in a similar way but reflects the \textit{actual} cost of providing electricity at a location given \textit{actual} system constraints. Despite the comparatively smaller volume, the real-time market plays a central role in determining the price of all futures contracts as the futures contracts' prices depends on the expected of the real-time market prices. The day-ahead market is a futures market that allows generators to enter agreements to provide electricity for the upcoming day.

Generators can fulfill obligations to provide energy by either physically producing electricity or purchasing it on the real-time market.
Multiple factors, such as unexpected maintenance, may cause a generator to fulfill their obligation through purchases on the real-time market rather than generation.
These factors, or risks, cause significant volatility in real-time markets compared to the day-ahead market \citep{FERC}.

Prices in the real-time market are a function of the cost to produce electricity and system constrains, such as congestion in transmission lines. 
When these constraints are binding, prices differ across locations in the PJM Interconnection to reflect the relative ease of delivering energy to a non-congested location and the relative difficulty of delivering energy to a congested location. Therefore, each node (or location) has an associated locational marginal price (LMP), which reflects the price of the marginal unit of electricity delivered to that specific location. LMPs are important price signals in the day-ahead and real-time market, which inform short-term decisions, as well as long-term investments and bilateral agreements \citep{cain2007common}.

\subsection{Local Marginal Price Data (LMP)}
Locational Marginal Pricing is used to price energy on the PJM market in response to changes in supply and demand and the hardware's physical constraints. LMP accounts for the cost to produce the energy, the cost to transmit this energy within PJM RTO, and the cost of energy lost due to resistance as the energy is transported across the system. LMP data is available at \textit{www.pjm.com}~\citep{PJM}. Our study uses price data, which includes real-time and day-ahead hourly prices from 1/1/2017 to 12/31/2017.
Load prices represent the cost of providing electricity at a given location. The price reflects the system's load (demand), generation, and limits of the transmission system.
The system's constraints can affect locations asymmetrically, causing variations in price across different locations.  Hub prices are a collection of these locational prices and are intended to reflect the uncongested price of electricity.

LMPs have three components: energy, congestion, and marginal loss. The energy component reflects the price of electricity, called system marginal price (SMP). SMP is calculated based on the current dispatch (supply) and load (demand). SMP is calculated for both the day ahead and real time markets. The congestion component is greater than zero whenever congestion occurs at a given node. Constraints occur when delivery limitations prevent the use of least-cost generator, for example, a higher cost generator located closer to load must be used to meet the demand if transmission constraints are present. The congestion price is calculated using the shadow price, which is the value of the dual variable (price of violating a binding constrain) in the optimization problem that governs the grid. When none of the constraints are active, all the congestion prices are zero.

The marginal loss component reflects the cost of transmission and other losses at a given location.  Losses are priced according to marginal loss factors, which are calculated at a bus and represent the percentage increase in system losses caused by a small increase in power injection or withdrawal.

\begin{table}[H]
\centering
\begin{tabular}{c|c}
	Variable name & Description \\\hline
	TotalLMP & Total cost; reflects Energy + congestion + marginal loss\\
	CongestionLMP & Congestion component of the LMP; can be $+$ve or $-$ve  \\
	& Value is relative to the energy component\\
	MarginalLMP & marginal loss component of the LMP\\
	\hline
\end{tabular}
\caption{Description of LMP variables.}
\end{table}

\subsection{Nonlinearities in Prices and Demand}      
\begin{figure}[H]
	\begin{tabular}{cc}
		\includegraphics[width=0.5\linewidth]{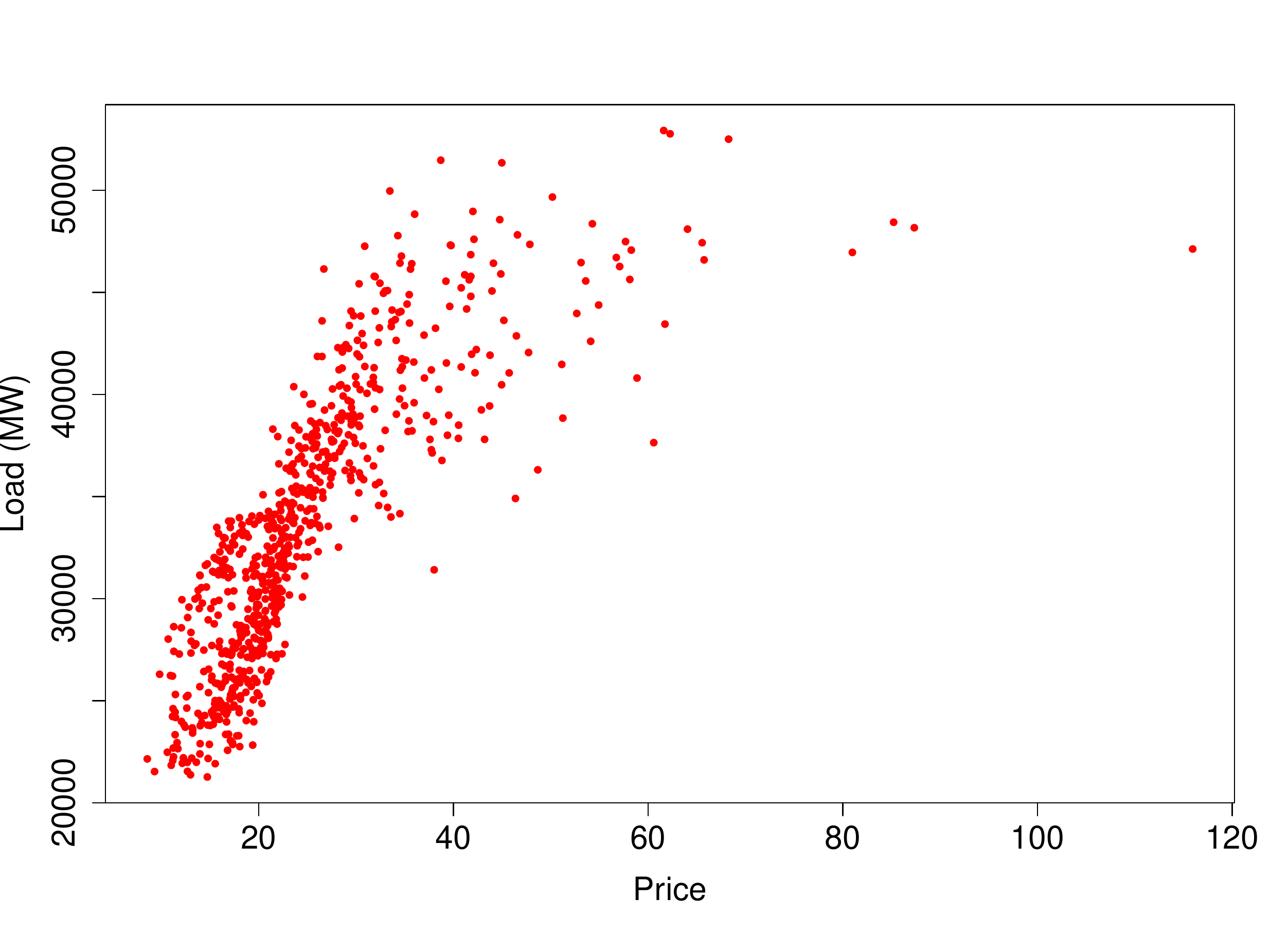} & \includegraphics[width=0.5\linewidth]{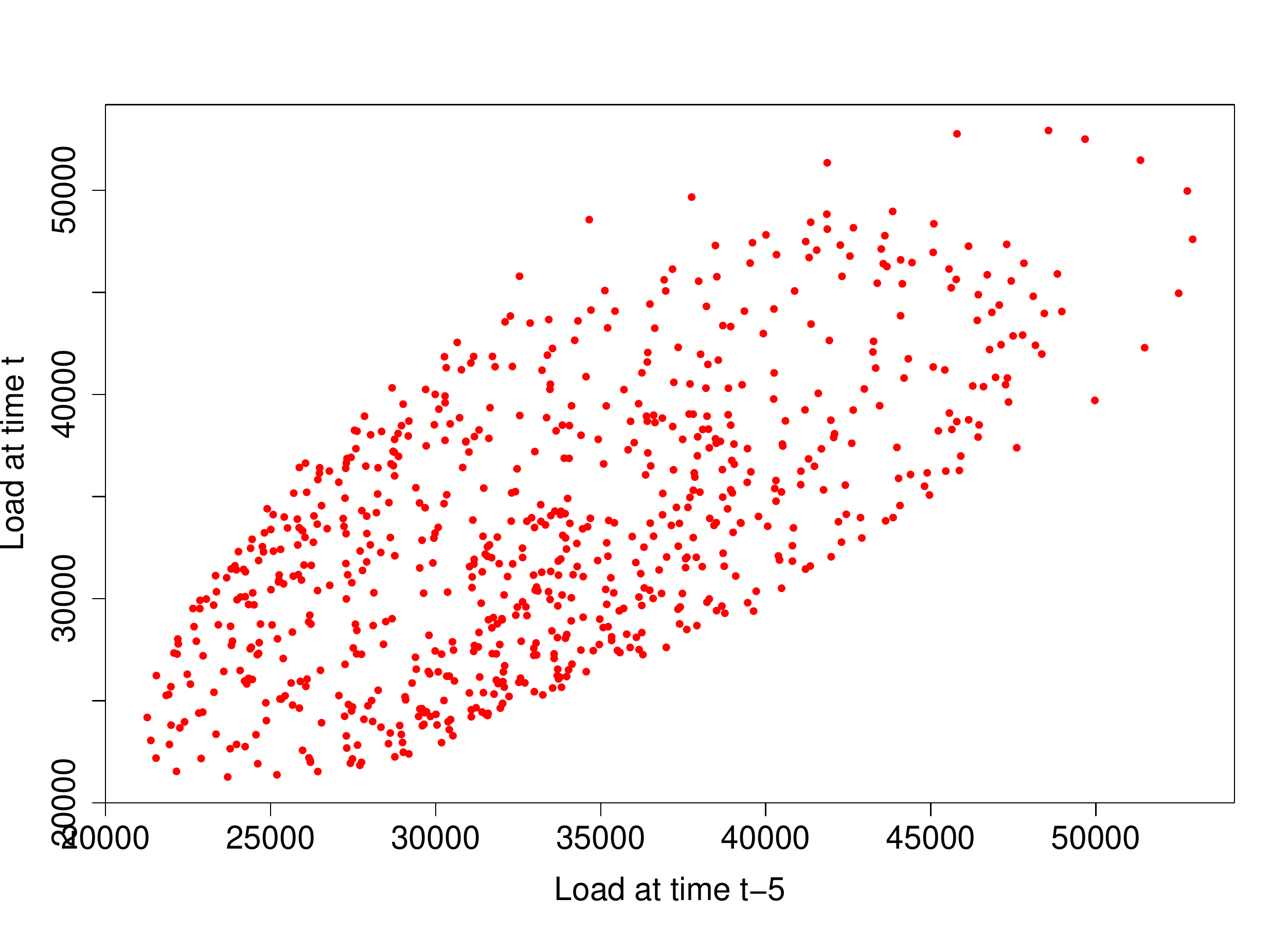}
	\end{tabular}
	\caption{Temporal patterns and nonlinearities in price-demand relationship}
	\label{fig:grid}
\end{figure}

The dynamics of energy prices are nonlinear due to the congestion component in the Locational Marginal Price. The congestion price represents the cost of violating a binding constraint in the linear program that models optimal generation strategy. The congestion price is paid by the load (consumer) to the generator (the producer). The congestion prices are calculated in real time and a day ahead. The constraints are the results of several physical limits of the electric grid and include thermal limits due to the power system equipment's thermal capability, voltage limits, and stability limits. Figure~\ref{fig:grid} shows the temporal patterns in the load data and the relationship between price and load variables. The figure shows that the relationships are non-linear.

The key to efficient electric grid management is understanding peak loads. At the day-to-day level, over- or underestimating peak loads can be costly. Overestimating peaks will cause the system to have too much generation in reserve. Underestimating peaks will cause the system to call upon costly, but flexible, sources of energy to quickly meet the demand. Day-to-day prediction is further complicated by the increase in renewable energy, whose pattern of generation does not always match the system's pattern of demand. This imbalance in supply and demand patterns adds to the volatility of the system's energy prices, further complicating prediction \citep{varaiya2011smart,hogg1983estimation}.

\section{Deep Learning}\label{sec:dl}
Let $y$ denote a low dimensional output and $x = (x_1,\dots,x_p)$ a high dimensional input. One of the advantages of DL models is the ability to analyze inputs with $p$ being in millions. For example, in image analysis 1 1 megapixel input image will correspond to $p = 10^{6}$.

DL uses a composite of univariate semi-affine rather than traditional additive functions. Deep learning models can efficiently approximate high-dimensional functions $y = F(x)$.
A deep network prediction, denoted by $ \hat y ( x )$, is defined by hierarchical layers 
\begin{align}
z_0 &  = x,~~z_1  = a_1 ( W_1 z_{0} + b_1 ),~\dots,~z_L  = a_L ( W_L z_{L-1} + b_L ) \nonumber\\
\hat y ( x ) & =  W_{L+1} z_{L} + b_{L+1}  \label{eq:dl}
\end{align}
where $W_l \in R^{n_l \times n_{l-1}}$ is the weight matrix, $b_l \in R$ is the bias term, and $n_l$ is the number of neurons in layer $l$. Here, we apply non-linear activation function $a_l$ element-wise to the activation vectors $(W_l z_{l-1} + b_l)$. Typical activation functions are rectified linear unit (ReLU) $a(x) =  \max (x, 0)$ and sigmoid $a(x) = 1/(1+e^{-x})$.

Specifically, the deep learning approach employs a series of hierarchical predictors comprising $L$ nonlinear transformations applied to the input $x$. Each of the $L$ transformations is referred to as a \emph{layer}, where the original input is $x$, the output of the first transformation is the first layer, and so on, with the output $\hat{y} $ as the last layer. Layers $1$ to $L$ are called \emph{hidden layers}. The number of layers, $L$, represents the \emph{depth} of our routine. Note that a linear regression is a particular case of a deep learning model with no hidden layers, e.g. $L=0$.

The fact that DL forms a universal ``basis'' which we recognize in this formulation dates to Poincare and Hilbert. From a practical perspective, given a large enough dataset of ``test cases",  we can empirically learn an optimal predictor. Similar to a classic basis decomposition, the deep approach uses univariate activation functions to decompose a high dimensional $x$.

It is well known that shallow networks are universal approximators and thus can be used to identify any input-output relations. The first result in this direction was obtained by \cite{kolmogorov57} who showed that any multivariate function can be exactly  represented using operations of addition and superposition on univariate functions. Formally, there exist continuous functions $\psi^{pq}$, defined on $[0,1]$ such that each continuous real function $F$ defined on $[0,1]^n$ is represented as
\[
F(x_1,\ldots,x_n) = \sum_{q=1}^{2n+1}a_q\left(\sum_{p=1}^{n}\psi^{pq}(x_p)\right),
\]
where each $a_q$ is a continuous real function. This representation is a generalization of earlier results~\cite{kolmogorov56,arnold1963}. \cite{kolmogorov56}  showed that every continuous multivariate function can be represented in the form of a finite superposition  of continuous functions of not more than three variables. 

Recurrent Neural Networks (RNNs) is a specific type of architecure designed to analyze sequences, e.g. time series data. RNNs can capture electricity prices' time series properties. Recurrent layers capture long term dependencies without much increase in the number of parameters. They learn temporal dynamics by mapping an input sequence to a  hidden state sequence and outputs via a recurrent relationships. Let $y_t$ denote the observed data and $h_t$ a hidden state, then
\begin{align*}
y_t &= \sigma(W_1h_t + b_z)\\
h_t &= \sigma(W_2[x_t,h_{t-1}] + b_h).
\end{align*} 
Here $\sigma(x) = 1/(1+ e^{-x})$ is the sigmoid function applied component-wise and is used for calculating both the hidden vector $h_t$ and the output vector $y_t$.
The main difference between RNNs and feed-forward deep learning is the use of a hidden layer with an auto-regressive component, here $h_{t-1}$. It leads to a network topology in which each layer represents a time step, and we index it by $t$ to highlight its temporal nature. 
\begin{figure}[H]
	\begin{center}
		\includegraphics[width=0.5\linewidth]{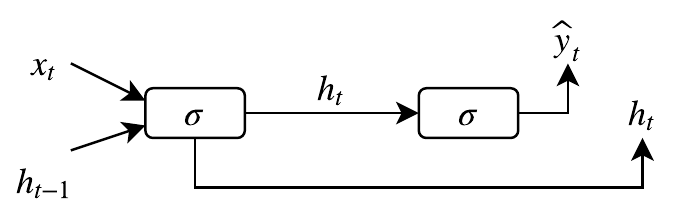}
	\end{center}
	\caption{Hidden layer of a Recurrent Neural Network}
	\label{fig:hl-rnn}
\end{figure}
A particular type of RNN, called LSTM (Long short-term memory), was proposed to address the issue of vanishing or exploding gradients in plain RNNs during training. A memory unit used in LSTM networks allows a network to learn which previous states can be forgotten \citep{hochreiter1997lstm,schmidhuber1997long}.  

\begin{figure}[H]
	\begin{center}
		\includegraphics[width=1\linewidth]{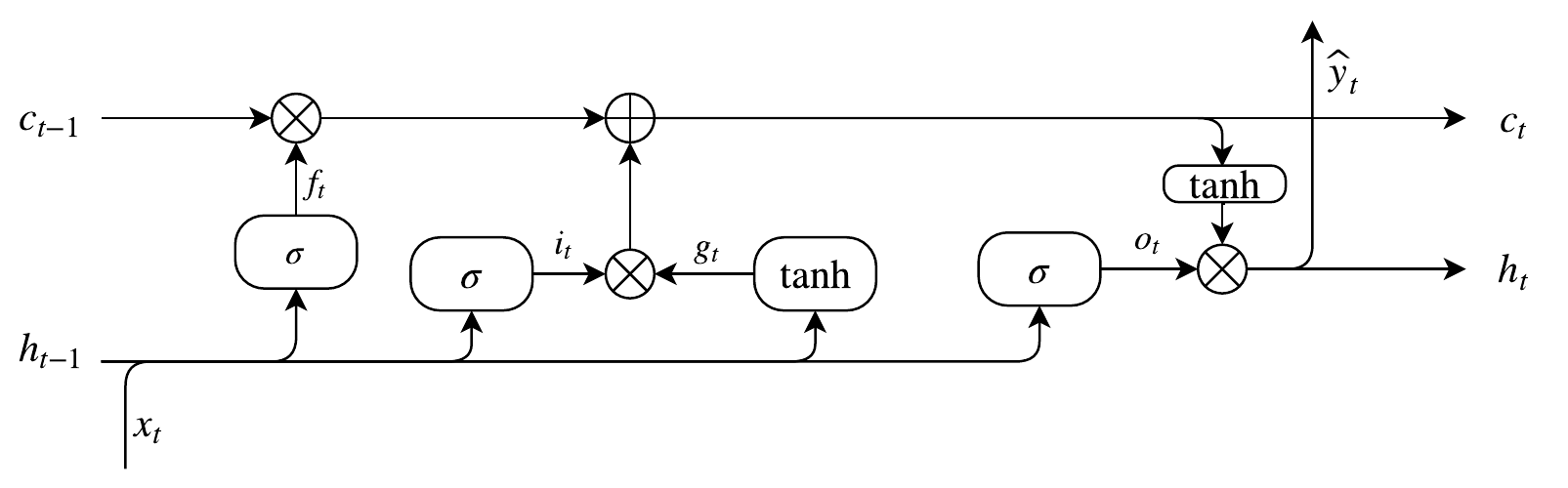}
	\end{center}
	\caption{Hidden layer of an LSTM model. Input $(h_{t-1},x_t)$ and state output $(h_t,c_t)$.}
	\label{fig:lstm}
\end{figure}

The hidden state will be generated via another hidden cell state $c_t$ that allows for long term dependencies to be ``remembered". Then, we generate
\begin{align*}
\text{Output: } & h_t = o_t \star \tanh(c_t)\\
& k_t = \tanh(W_c[h_{t-1},x_t] + b_c)\\
&c_t = f_t \star c_{t-1} + i_t \star k_t\\
\text{State equaitons: }& \left(\begin{array}{c}
f_t\\i_t\\o_t
\end{array}\right) = \sigma(W[h_{t-1},x_t] + b).\\
\end{align*}
Where $\star$ denotes point-wise multiplication. Then, $f_t \star c_{t-1}$ introduces the long-range dependence. The states $(i_t,f_t, o_t)$ are input, forget, and output states.  Figure \ref{fig:lstm} shows the network architecture.



The key addition, versus RNN is the cell state $c_t$, the information is added or removed from the memory state via gates defined via the activation function $\sigma(x)$ and point-wise multiplication $\star$. The first gate $ f_t \star c_{t-1}$, called the forget gate, allows to throw away some data from the previous cell state. The next gate $ i_t \star k_t$, called the input  gate, decides  which values will be updated. Then, the new cell state $c_t$ is a sum of the previous cell state $c_{t-1}$, passed through the forgot gate plus selected components of the $k_t$ vector, which is a filtered version of inputs $(h_{t-1},x_t)$. Thus, the vector $c_t$ provides a mechanism for dropping irrelevant information from the past, and adding relevant information from the current time step. At the last output layer, the filtered version of of the previous hidden state and input vectors $o_t$ is then combined with $\tanh$ applied to the cell state $o_t \star \tanh(c_t)$. The forget gate resolves the problem of vanishing gradient, which is the case when values of the gradient vector are close to zero. SGD optimization algorithm is straight forward to implement. See Section~\ref{sec:sgd} for discussion. 

Deep rectified linear units (ReLU) with long short term memory (LSTM) cells have became popular architectures as they can capture long-range dependencies and  nonlinearities. They can efficiently approximate highly multivariate functions with small number of neurons at each layer \citep{bach2017breaking,schmidt2017nonparametric, yarotsky2017error}.

\section{Deep Learning Extreme Value Theory (DL-EVT)}\label{sec:evt}
A traditional deep learning regression model uses least squared loss to estimate model parameters (weights and biases of each of the neural network layers). 
This model is not appropriate for quantifying large values of $y$ (spikes) that are a rare but very crucial to the stable operations of electric grids. Extreme value theory approach allows to model the tail behavior of the distribution of electricity loads. 

Let each observation follows a common distribution $y_i \sim G(y_i) = \mathrm{Pr}(Y \le y_i)$ and let $M_n = \max\{y_1,\ldots,y_n\}$. The central result of the extreme value theory is that regardless of the distribution $G$, the scaled value of $M_n$ follows a limiting distribution $K$
\[
\mathrm{Pr}\left\{ \frac{M_n-b_n}{a_n}\le y \right\} = G^n(a_ny+b_n)\rightarrow K(y)
\]
Here $a_n >0$ and $b_n$ are normalizing constants. \cite{gnedenko1943distribution} provided a rigorous mathematical proof of existence of this limiting distribution and characterized its functional form \citep{davison1990models}. Modeling the extreme values $M_n$ limits the number of samples that can be used for estimation. For example, we can use monthly maximum loads but then we will have to discard most of the samples. Further, traditional extreme value theory does not allow for covariates (predictor inputs). It makes this approach impractical. A more practical approach was suggested by \cite{smith1989extreme} who proposed to model values of $y$ that exceed some fixed threshold value $u$. The distribution over the excess values has a limiting distribution as $u+y$ approaches the right-hand endpoint of the underlying distribution. Specifically, as $u+y$ approaches the right-hand side endpoint of distribution $G$, for some normalizing constant $c_n >0$, we have
\[
\mathrm{Pr}\left\{ Y \le c_u(u + y) \mid Y > u  \right\} \rightarrow H(y)
\]
where 
\[
H(y\mid \sigma,\xi) = 1-\left(1+\xi\dfrac{y-u}{\sigma}\right)^{-1/\xi}_+,~~~\xi\ne 0.
\]
Distribution $H(y)$ is  called the generalized Pareto (GP) distribution. The corresponding density function is given by
\[
h(y \mid \sigma,\xi) = \dfrac{1}{\sigma}\left(1+\xi\dfrac{y-u}{\sigma}\right)^{-1/\xi-1},~~~1+\xi\dfrac{y-u}{\sigma}>0,~\xi \ne 0
\]
Here  $(u,\sigma,\xi)$ are the location, scale and shape parameters,  $\sigma > 0$ and $z_+ = \max(z, 0)$. The Exponential distribution is obtained by continuity as $\xi \rightarrow 0$, and we have
\[
\lim\limits_{\xi \rightarrow 0} h(y \mid \sigma,\xi) = \sigma\exp\left(-\sigma(y-u)\right)
\]
Under this distribution, the mean value of the $y$ is $\sigma + u$.

We fit the GP distribution using observations that exceed the threshold. Compared to the the the classical extreme value theory that only models maximum values, each exceedance is associated with a specific time point and it allows us to incorporate covariates, e.g. when parameters $\sigma$ and $\xi$ depend on input variable $x$.

Suppose, that we have data denoted by $y(s_i,t_j)$ at spatial locations $s_i$, $1 \le i \le n$ and time $t_j$, $1 \le j \le T$ and. We build an EVD-DL input-output model for each location $s_i$ and estimate it using the pairs $\{(y(s_i,t_k), x_k)\}_{k\in C}$, where $C = \{j \mid Y(s_i,t_j)>u\}$ and $x_k = (y(s_i,t_k), y(s_i,t_{k-1}),\dots,y(s_i,t_{k-h}))$ are the recent observations of the output for the given location. 

We assume that observations follow GP distribution with parameters being functions of the input variables
\[
	y(s_i,t_k) \mid x_k \sim \mathrm{GP}(\sigma(x_k),\xi(x_k))
\]

We model the functions $\sigma(x\mid W,b)$  and $\xi(x\mid W,b)$ using a deep learning model parametrized by weight matrices $W = (W_1,\dots,W_{L+1})$ and biases $b = (b_1,\dots,b_{L+1})$.  Linear regression GP  were developed in ~\citep{davison1990models,beirlant2006statistics}.  To complete our specification for exceedance sizes we assume a functional form for $\sigma(x\mid W,b)$ that is a deep neural network. As shown in Equation \ref{eq:dl}, we introduce 
\[
(\xi(x),\sigma(x)) = F\left(x\mid W,b\right), \text{ where } F = f_l \circ \ldots \circ f_L, ~~f_l(z) = \sigma(W_lz + b_l)
\]
Here $F$ is a deep learner constructed via superposition of semi-affine univariate functions, see~\cite{polson2017deep,polson2015bayesian,dixon2017deep} for further discussion. 

To estimate the weights and bias parameters of the deep leanring model, we use the the negative log-likelihood loss function. Under, the the assumption of GP distribution for our dependent variable, for a single observation, the negative log-likelihood is given by
\begin{align*}
	l(y_i\mid x_i,W,b) = & \log \sigma(x_i \mid W,b) \\
	 & -(1/\xi(x_i \mid W,b) +1)\log\left(1+\sigma(x_i \mid W,b)\xi((x_i \mid W,b))(y_i-u)\right).	
\end{align*}


Then, the loss function for our deep learning model model, which is the negative log-likelihood for a training data sample,  becomes
\[
L(W,b) = \dfrac{1}{|C|}\sum_{i\in C} l(y_i\mid x_i,W,b) .
\]
The weights $W$ and offsets $b$ are learned by minimizing the loss function, using stochastic gradient descent (SGD) algorithm.


%
%
%
%

\section{Empirical Results}\label{sec:application}
In this section we compare temporal neural network architecture with more traditional Fourier and ARIMA models to predict electricity prices. Further, we demonstrate our DL-EVT approach to predict peak loads on the PJM interconnect. 

\subsection{PJM Price Forecasting}
\begin{figure}[H]
	\centering
	\includegraphics[width=0.5\linewidth]{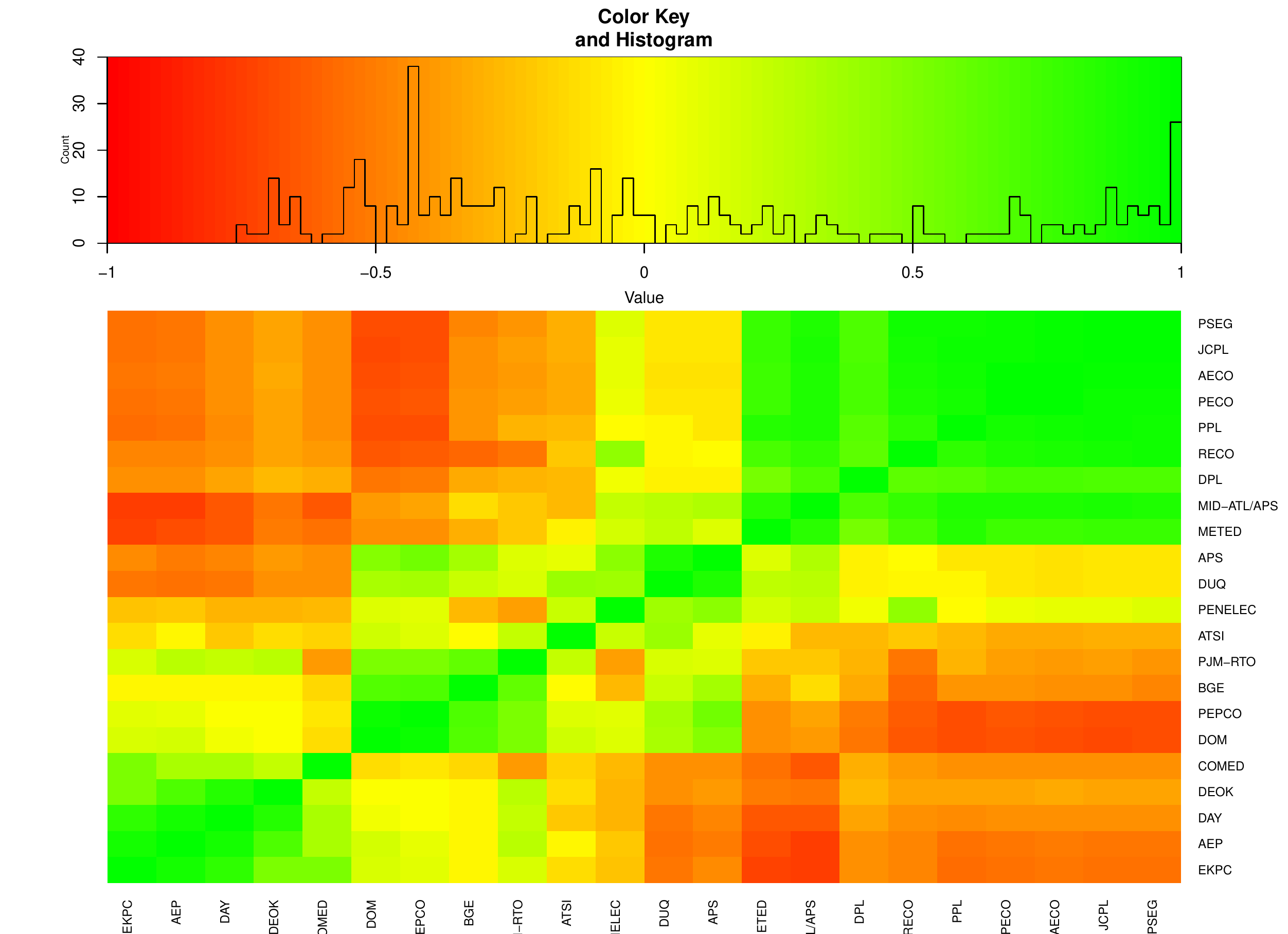}
	\caption{Correlation in Marginal Prices Among Zones}
	\label{fig:zone_proce_cor}
\end{figure}

We predict the price at each load point using the historical prices, demand and weather observations. There are a total of 4719 generating nodes in the system. Plot~\ref{fig:zone_proce_cor} shows that there are strong spatial correlation among prices at different zones. Zone aggregates several nodes. Thus, prices at nodes will be correlated as well. We use this spatial pattern to build our model. 

\begin{figure}[H]
	\begin{tabular}{cc}
		\includegraphics[width=0.5\textwidth]{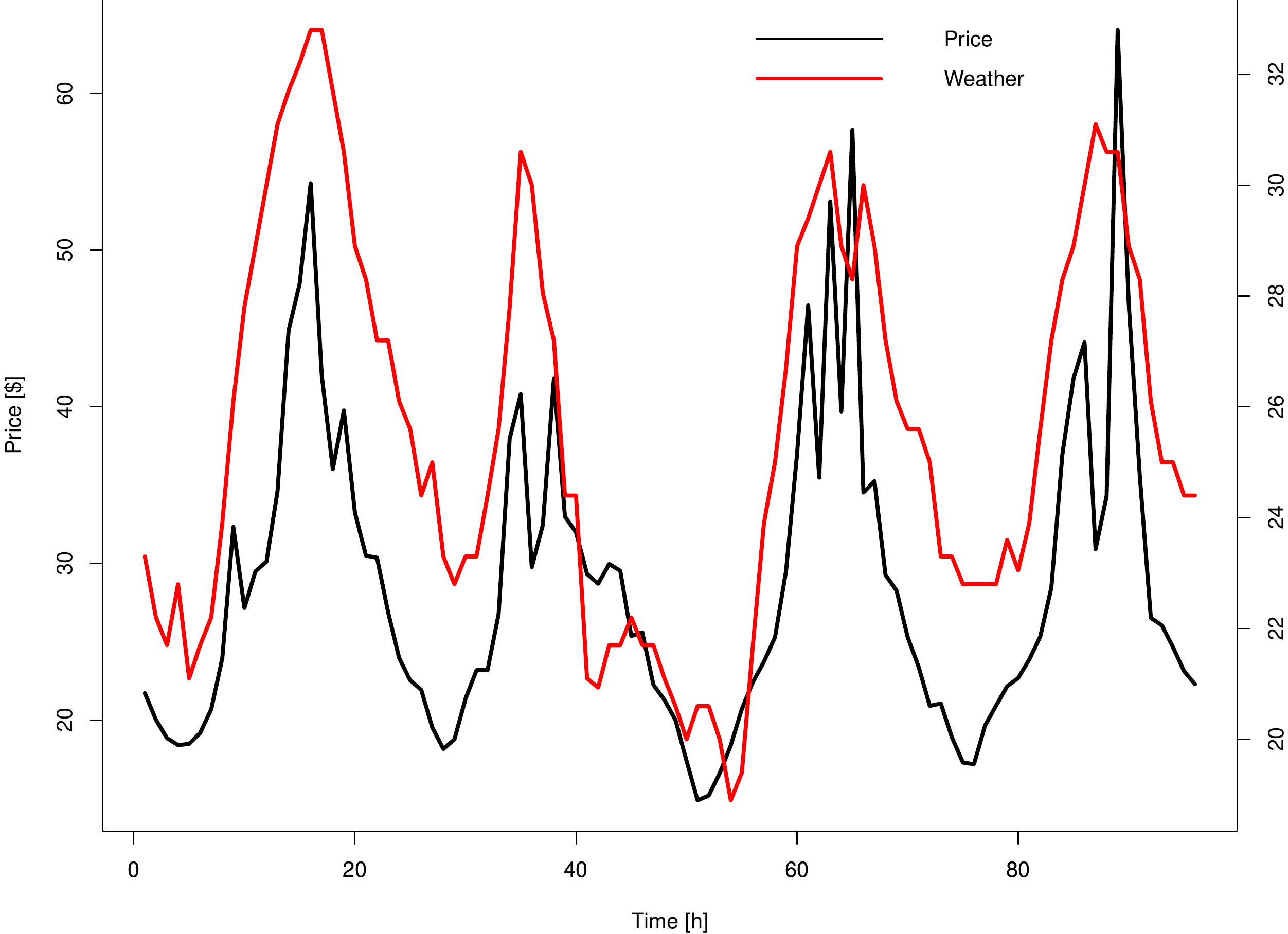} & \includegraphics[width=0.5\textwidth]{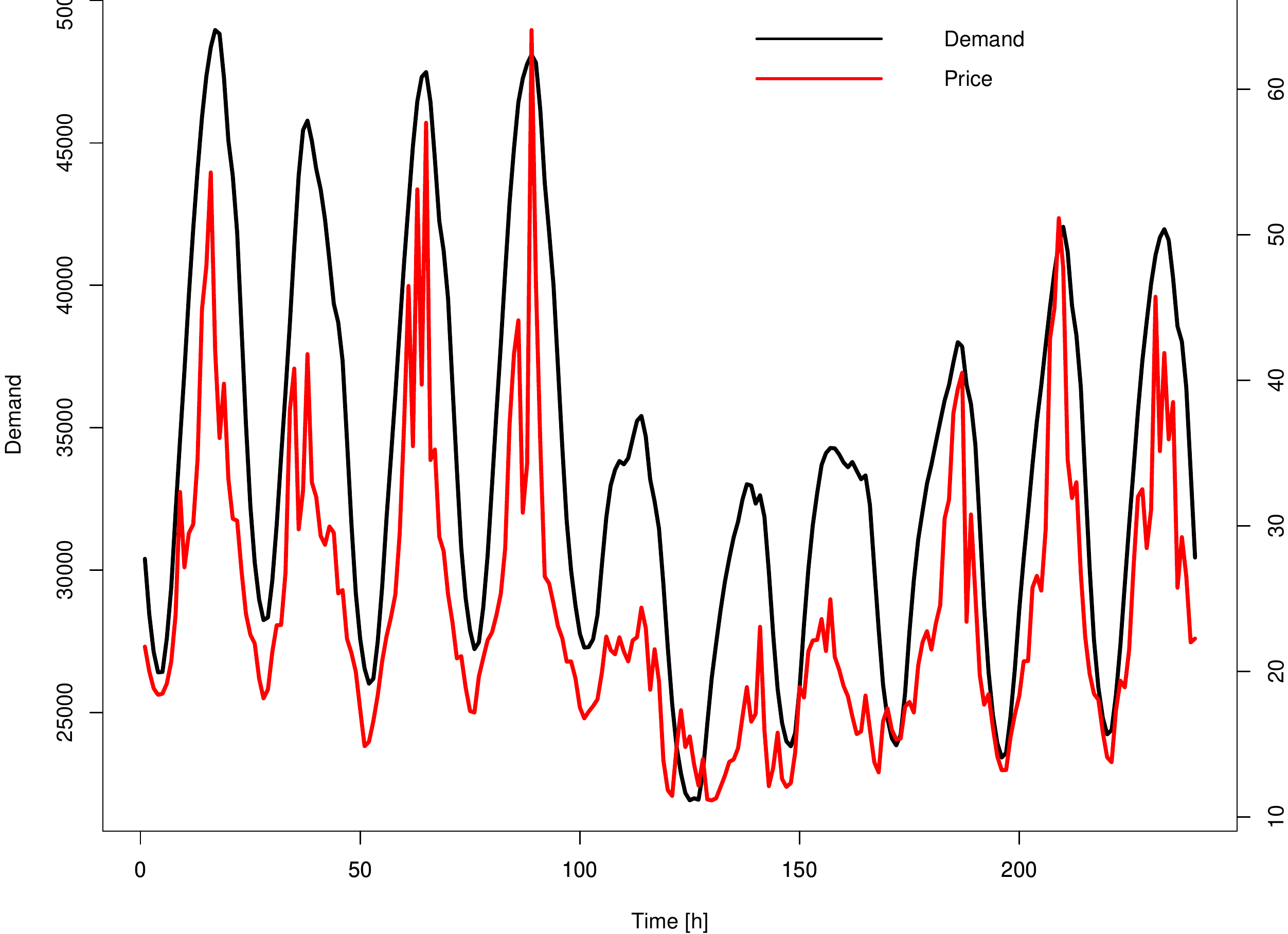}\\
		(a) & (b) 
	\end{tabular}
	\caption{(a) Price vs Temperature [C]. (b) Demand vs Price}
	\label{fig:covariates}
\end{figure}

Further, Figure~\ref{fig:covariates} shows the relations between price and weather as well as price and the demand. We see that demand is not alway an accurate predictor for the price. This is due to the nonlinearities present in the system. High demand not necessarily leads to high prices. We only observe demand variable for the overall system and not on individual nodes. Thus, we cannot use spatial patterns in the demand variable to predict the prices. 

We demonstrate our forecasting approach for a node named "KULLERRD138 KV  T-2", with id 48667, which is locate in Clifton, NJ.

First, we try traditional model for electricity prices, which  uses Fourier series to describe the seasonal patterns and short-term time series dynamics modeled as an ARIMA terms. Here $y_t$ is decomposed as a sum, a deterministic Fourier term $f(t)$, and a stochastic component, $N_t$, leading to
\[
y_t = a+ f(t) + N_t,
\quad\text{where}\quad
f(t) =  \sum_{k=1}^K \left[ \alpha_k\sin(2\pi kt/m) + \beta_k\cos(2\pi kt/m)\right],
\]
where $N_t$ is an ARIMA process. The number of terms of $K$ can be chosen by minimizing cross-validation. This allows: (i) any length seasonality, (ii) several seasonality periods. Smoothness of the seasonal term  is governed by value $K$.
The short-term dynamics is handled with an ARMA error. The only real disadvantage (compared to a seasonal ARIMA model) is that the seasonality is assumed to be fixed — the pattern is not allowed to change over time. In practice, seasonality is usually remarkably constant, so assumption generally holds except in applications with very long time series.

\begin{figure}[H]
	\includegraphics[width=1\textwidth]{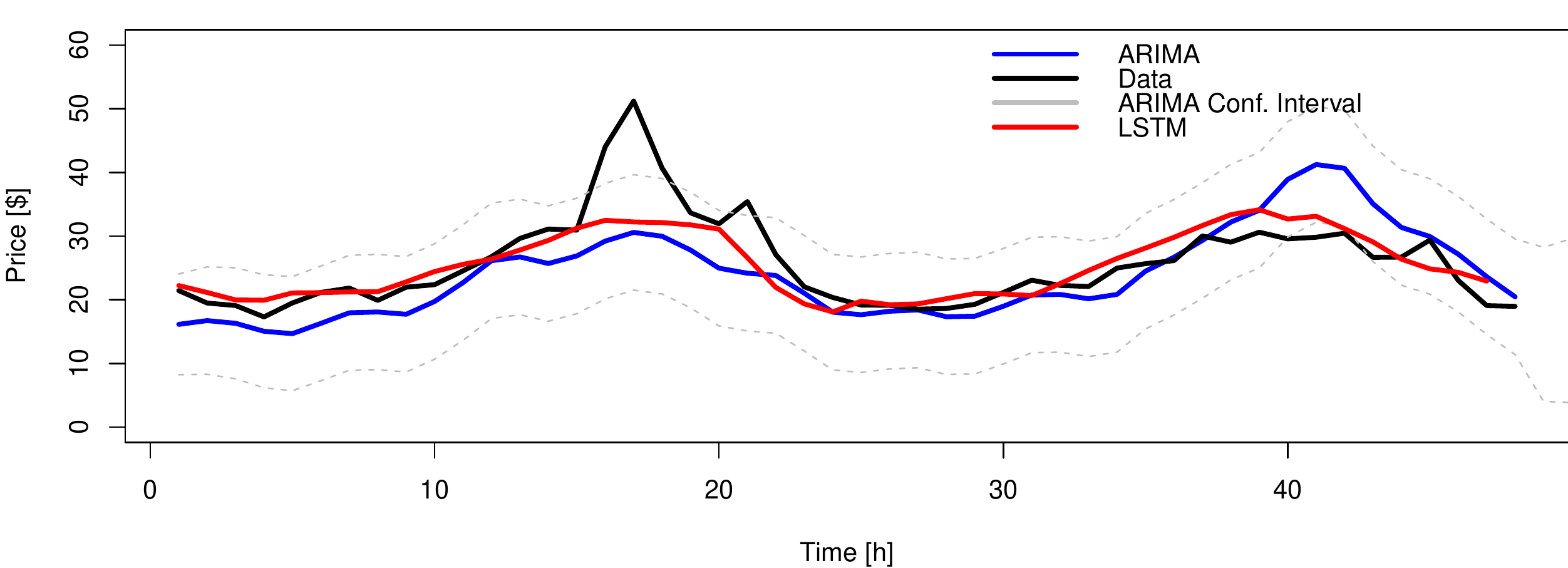}
	\caption{Comparison of Fourier and DL models}
	\label{fig:compare}
\end{figure}

The in-sample fit of the Fourier model is shown on Figure~\ref{fig:compare}.
\begin{figure}[H]
	\centering
	\includegraphics[width=0.6\textwidth]{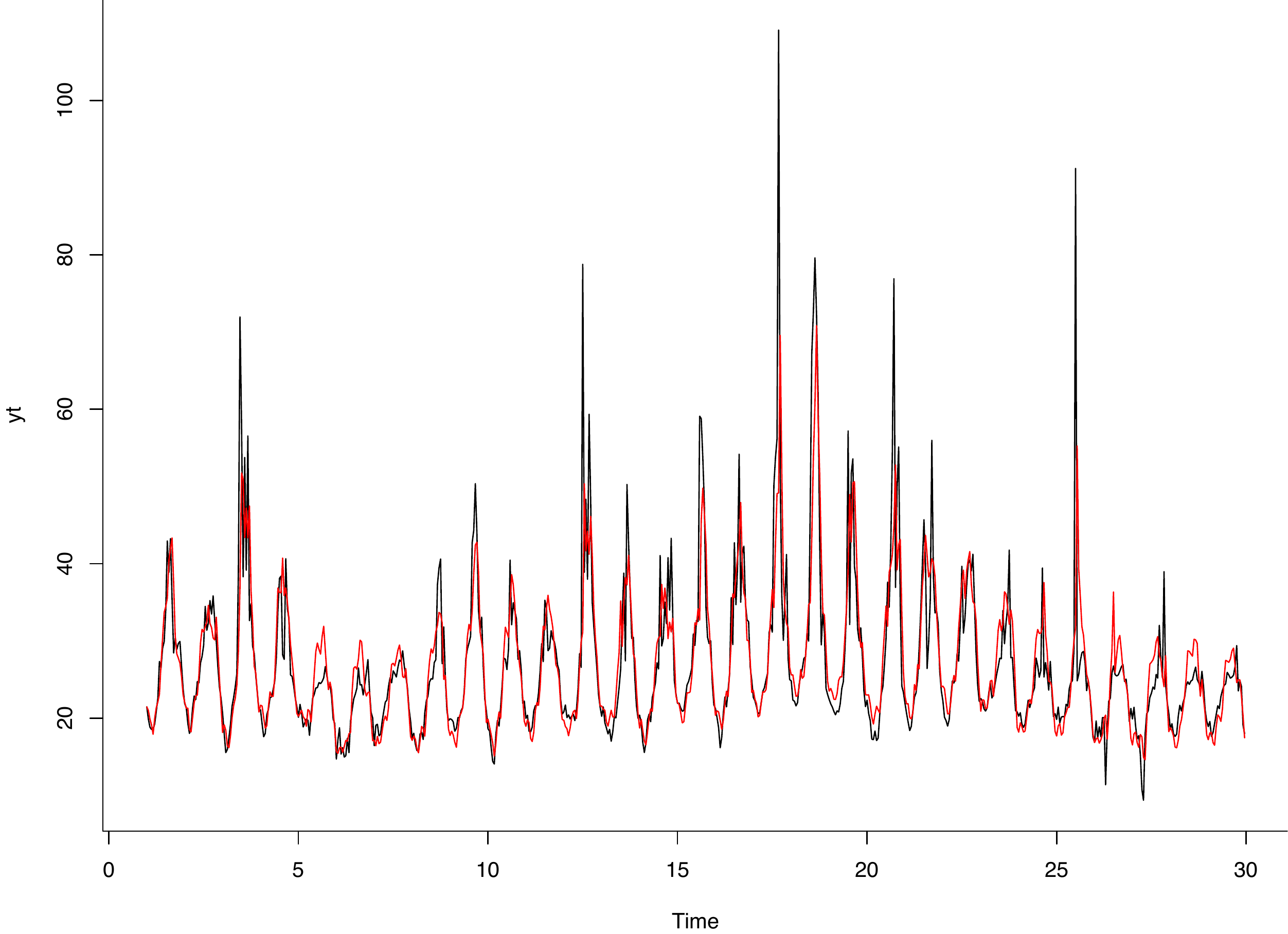}
	\caption{In-sample prediction}
	\label{fig:arima}
\end{figure}
This model captures the cyclical patterns in the prices but does not accurately capture the levels of the peak prices. 

\begin{figure}[H]
	\begin{tabular}{cc}
		\includegraphics[width=0.5\textwidth]{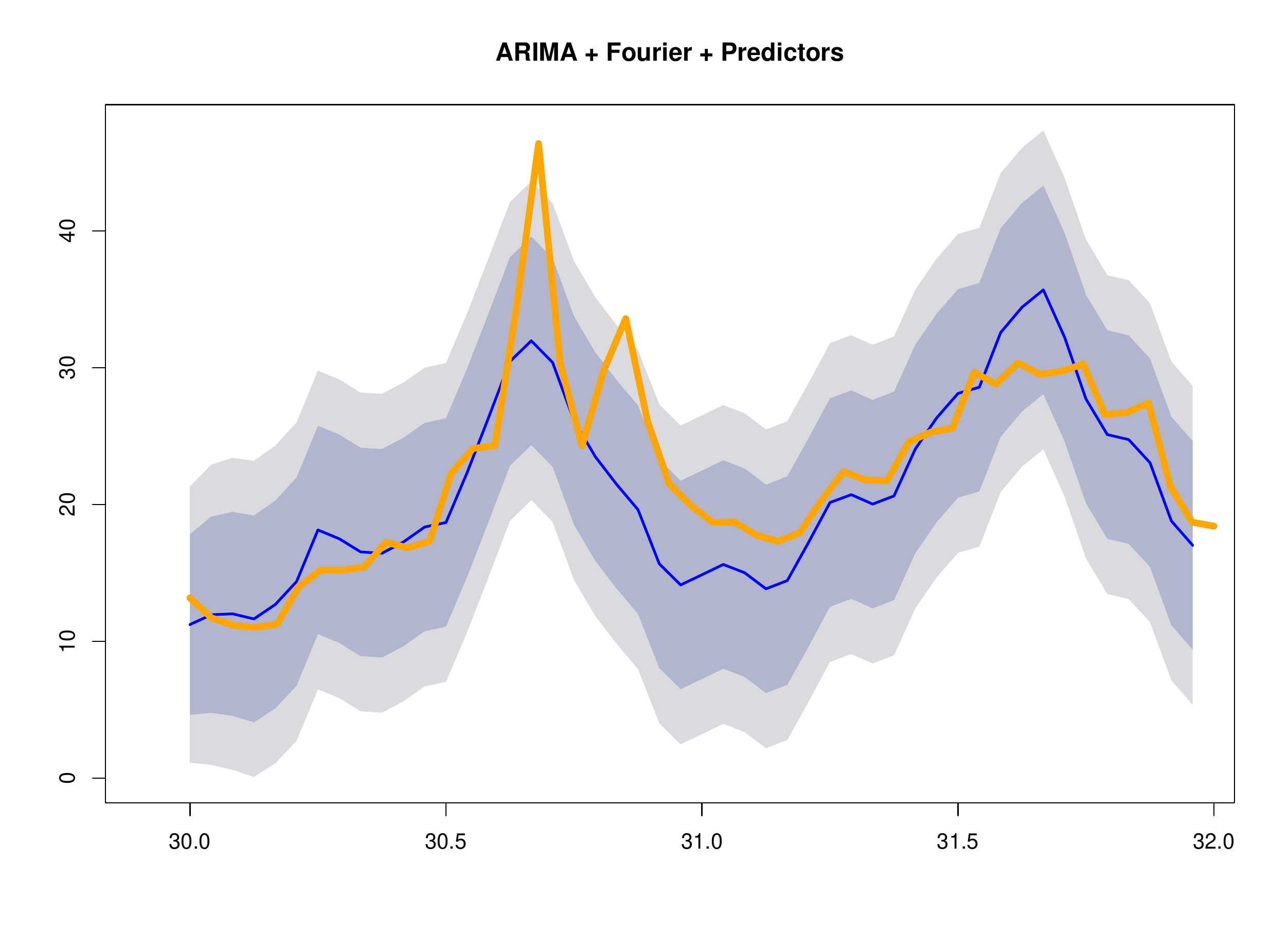} & \includegraphics[width=0.5\textwidth]{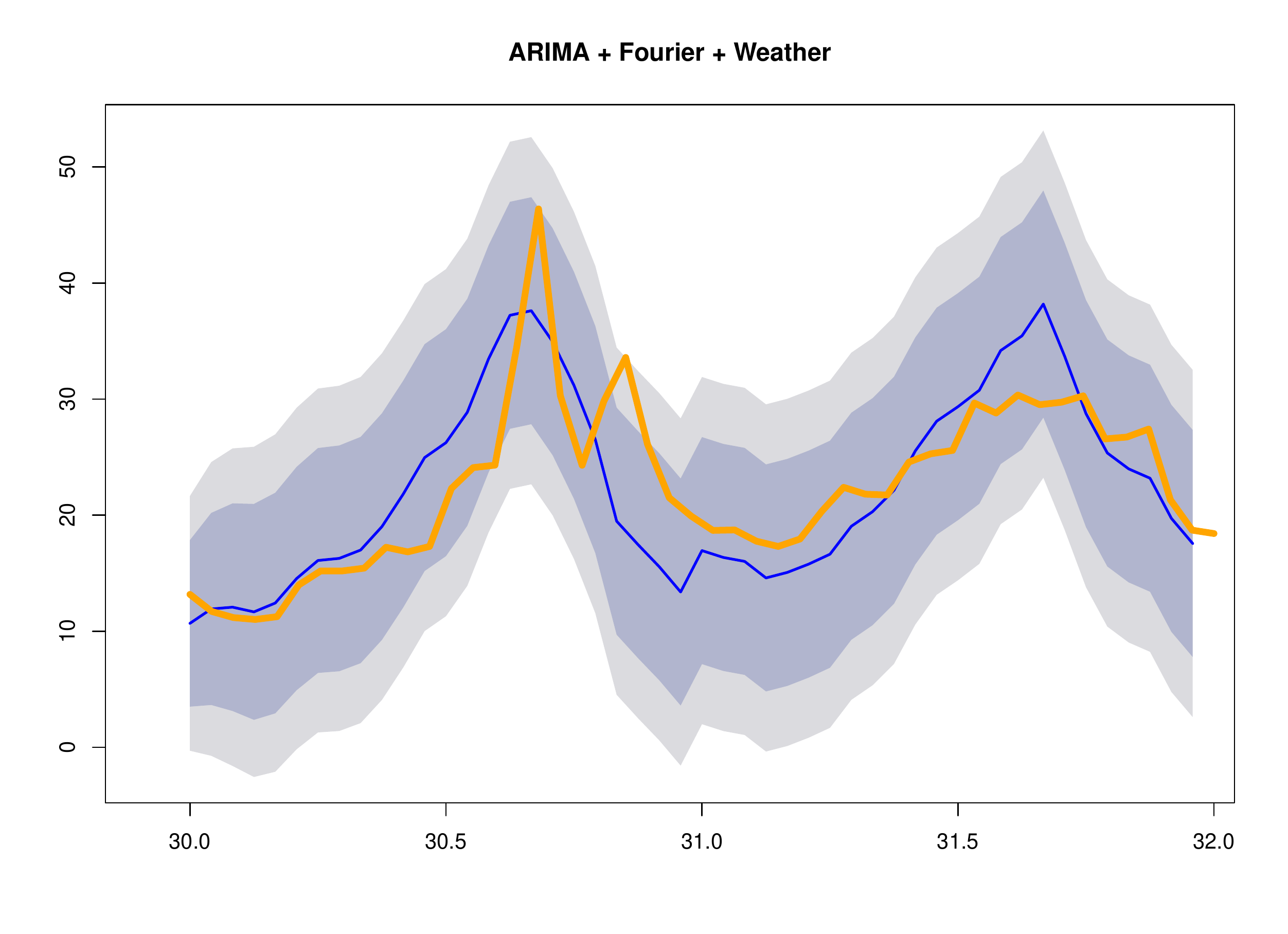}\\
		(a) Demand + Weather & (b) Only weather
	\end{tabular}
	\caption{Out-of-sample Prediction from linear model with ARIMA$_{(2,0,0)}$ errors and Fourier predictors. Yellow is data and blue is the forecast with confidence intervals.}
	\label{fig:arimaos}
\end{figure}

Figure~\ref{fig:arimaos} shows the out-of-sample prediction for the 32 hours of price observations for Fourier model with weather and demand predictors. Inclusion of predictors does not change the quality of forecasting peak prices. As we noted in our exploratory plots, demand is not a good predictor of a peak price. 

We compare the Furies model with temporal neural network (LSTM) model. Table~\ref{tab:fourier-lstm} shows several out-of-sample goodness-of fit metrics.
\begin{table}[H]
\centering
\begin{tabular}{c|c|c|c|c}
	& mse & mrse & mae & mape \\ 
	\hline 
	ARIMA + Fourier    & 26.6 & 5.1 & 4 & 0.19 \\ 
	LSTM& 16.8 & 4.1 & 2.4 & 0.09 \\ 
\end{tabular} 
\caption{Out-of-sample performance of DL and Fourier models}
\label{tab:fourier-lstm}
\end{table}


LSTM model shows an improved performance when compared to traditional ARIMA model with Fourier predictors. However, as shown in Figure~\ref{fig:compare} both, the traditional ARIMA and LSTM neural network model are not able to capture the peak in the price value at 5pm. Further, the peak price lies outside of the 95\% confidence interval of our ARIMA model. On the other hand, prediction of the peak values is of high importance. In the next section we show how EVT combined with DL (DL-EVT) addresses this problem and captures the peak values of the demand time series.  
%

%
%
%
%
%
%
%
%
%
%
%
%
%
%

\subsection{Demand Forecasting}
Electricity load forecasting is essential for designing operational strategies for electric grids. In presence of renewable energy sources short-term forecasts are becoming increasingly important. Many decisions, such as dispatch scheduling and demand management strategies are based on load forecasts~\cite{taylor2007short}. One hour-ahead forecasts are a key input for transmission companies on a self-dispatching markets~\cite{garcia2006forecasting}. Hourly behavior of electricity load is known to be non-stationary~\cite{almeshaiei2011methodology}. Since there is not much of a change in meteorological variables, it is typical to use univariate time series data for short term load forecasting~\cite{bunn1982short}.

In this section, we analyze an hourly electricity load observations on the PJM interconnection. The data is available  at \url{https://www.dropbox.com/sh/1dczb673bx9kxzl/AABII5ePMWdFhAk-dEcRGS1La?dl=0}
We use data for January 2016 - May 2018 period  year of observations and use last 10 days of observations for testing.  We use a local trend model that takes previous 24 observations (one day) to predict load in  five hours.

\begin{figure}[H]
	\begin{tabular}{cc}
		\includegraphics[width=0.5\linewidth]{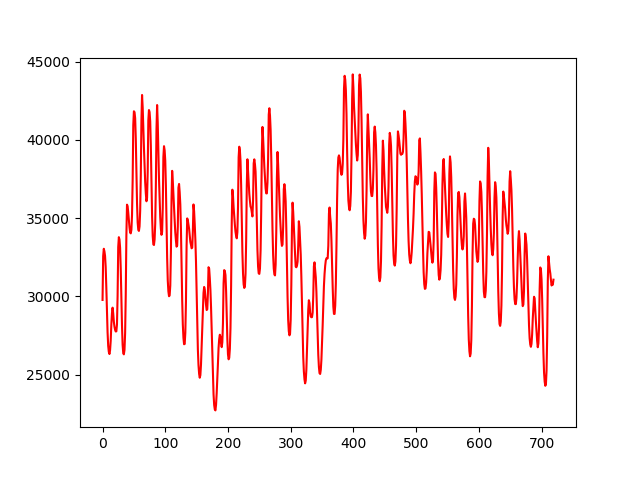} & \includegraphics[width=0.5\linewidth]{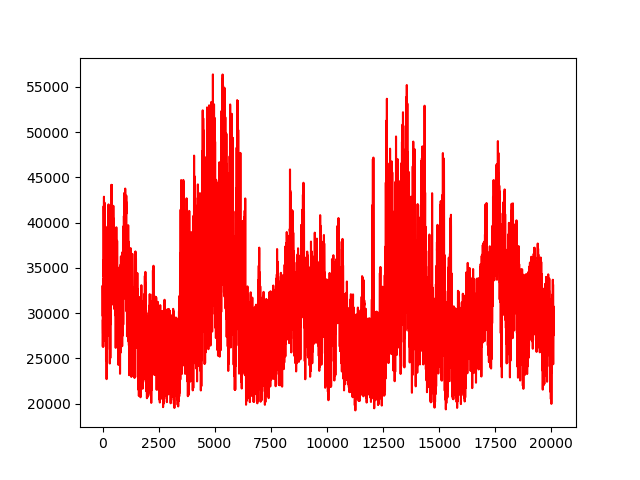}\\
		(a) Jan-Feb 2016 & (b) Jan 2016 - Apr 2018
	\end{tabular}
	\caption{Hourly electricity load on PJM interconnect in MW.}
	\label{fig:loadtrain}
\end{figure}

Figure~\ref{fig:loadtrain}(b) shows the hourly PJM interconnect load series from January 2016 to April 17, 2018. This data was used to train our DL-EVT model. The graph shows seasonal cycles. Figure~\ref{fig:loadtrain}(a)  shows the shorter period (Jan-Feb 2016) of the same data and shows daily and weekly cycles. We can see that weekends have lower load levels compared to work days.	

The following architecture is used to model the relations between previous load observations ($X$) and the scale parameter of the Generalize Pareto distribution $\sigma$.

\includegraphics[width=0.9\linewidth]{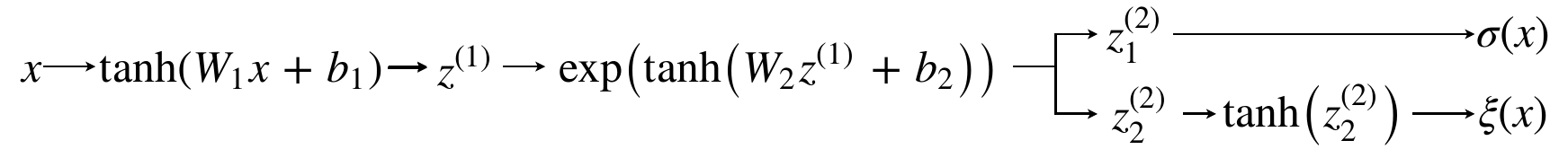}
	
\noindent where $W_1\in R^{p\times 3}$ and $W_2 \in R^{2\times 3}$, and $x\in R^{p}$ is the vector of recent observations of electricity demand, we used $p = 24$ (1 day). We use $\tanh$ to constrain values of $\xi$ to be in the (-1,1) interval. We require $\xi<1$ to guarantee that we have a finite mean, which we use in our prediction. Further, we require $\xi>-1$ to guarantee that the likelihood function is bounded.

\begin{figure}[H]
\begin{tabular}{cc}
\includegraphics[width=0.5\linewidth]{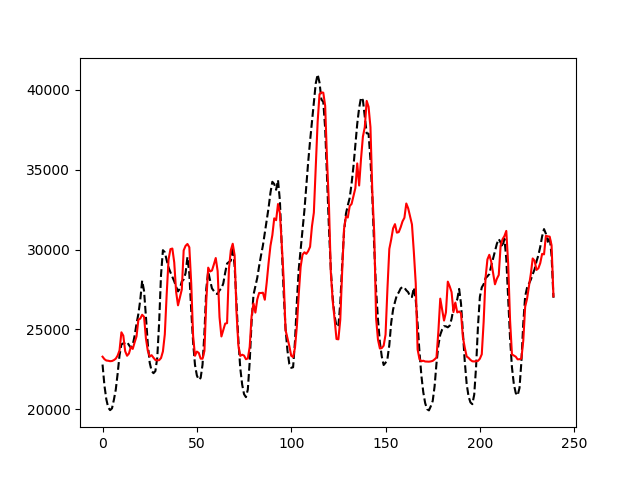} & \includegraphics[width=0.5\linewidth]{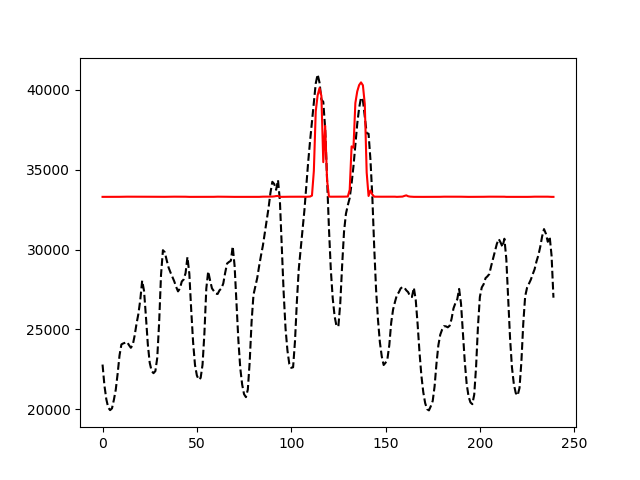}\\
(a) DL (MSE Loss) & (b) DL-EVT (GP Loss)
\end{tabular}
\caption{One our electricity load and its forecast for the period from Friday, April 27, 2018 to Monday, May 7, 2018}
\label{fig:load_dl}
\end{figure}

To train the EVT model we only used the observations $y_i > u$ with $u = 31000$. We used the mean $\sigma/(1-\xi) + u$ of the GP distribution as the point estimate for plotting Figure~\ref{fig:load_dl}(b).

Our DL-EVT model is compared with a vanilla deep learning model with standard mean squared error loss function. Figure~\ref{fig:load_dl} shows the resulting out-of-sample forecasts.
We can see that while a standard DL model captures both ups and downs in the load levels, the DL-EVT model does capture the location and level of the peak loads more accurately compared to the standard DL model.

\section{Discussion}\label{sec:discussion}
Deep learning, combined with extreme value theory, can predict peaks in electricity prices and demand. With the availability of real-time data, computational power, and machine learning pattern recognition tools, such as deep learning, we have the ability to more accurately predict and manage energy generation and distribution. One of our goals is to demonstrate that an EVT extension of the standard DL framework is a viable option and is applicable to electricity data. DL-EVT performed well on in- and out-of-sample forecasting of electricity prices and load. 

We demonstrated our DL-EVT model is more accurate at forecasting peak values that exceed a given threshold when compared to a Gaussian likelihood-based model. The EVT model predicts peak values conditional on the exceedance over the threshold. One of the artifacts of our model is that prediction for time points when a threshold is not expected to be exceeded is a constant value. An extension of our approach could include a binary classifier that predicts the probability of crossing a specific threshold. Another possible extension is to include a Gaussian likelihood-based model to forecast values below a threshold. ~\cite{naveau2016modeling} demonstrate that a similar approach can be used for successful environmental modeling.

Forecasting electricity prices is challenging because they can spike due to supply-demand imbalances, yet have long-range dependence. Deep ReLU LSTM models capture spikes with non-linear activation functions, are scalable, and can efficiently fit using SGD. For a grid of 4,786 electricity load nodes, we show how such models can fit in-sample with better accuracy than traditional time series models. 
There are a number for directions of future research. For extensions to multivariate time series data with spatiotemporal dynamics, see~\cite{dixon2017deep}.

\bibliographystyle{plainnat}
\bibliography{ref}

\appendix
\section{Stochastic Gradient Descent for Deep Learning (SGD)}\label{sec:sgd}
Once the activation functions, depth $L$ and size $n_1,\ldots,n_L$ of the learner have been chosen, the parameters, $\hat{W} $ and $\hat{b} $ are found by solving the following optimization problem
\begin{equation}
\operatorname{minimize}\limits_{W, b} \sum_{i=1}^T L( y_i , \hat{y} ( x_i \mid W,b))+ \phi (W ), 
\label{Training_Eq1}
\end{equation}
Which is a penalized loss function, where $(y_i,x_i)_{i=1}^N$ is training data of input-output pairs, and $ \phi( W ) $ is a regularization penalty on the network weights. Most architectures employ regularization techniques to prevent the model from over-fitting training set data \citep{Hinton}. This improves the model's predictive performance on data outside of the training set. Normally, a regularization penalty allows to improve convergence rate of the optimization algorithms and to avoid over-fitting. Dropout, the technique of removing input dimensions in $x$ randomly with probability $p$, can also be used to further reduce the change of over-fitting during the training process \citep{Srivastava}.

A typical choice is $L(y_i , \hat{Y}( x_i))$ being an $L_2$-norm, then we have a traditional least squares problem \citep{Janocha} and $\phi(W) = \lambda ||W||_2$.

The common numerical approach to find the solution to this optimizaiton probelm \ref{Training_Eq1} is stochastic gradient descent (SGD). It iteratively updates the current iterated by taking a step in the direction opposite to the gradient vector
\[
(W,b)^+ = (W,b) - \eta \nabla \left(L( y_i , \hat{y} ( x_i \mid W,b))+ \phi (W )\right)
\]
SGD then uses back-propagation algorithm to calculate the gradient at each iteration. Back-propagation is an implementaiton of chain rule applied to a function defined by a neural network. One caveat of SGD is the complexity of the system to be optimized, resulting in slow convergence rates. As a result, deep learning methods rely heavily on large computational power \citep{abadi2016tensorflow,Cardoso}.

\end{document}